\documentclass{article}
\usepackage{sty/ijcai22}

\usepackage{times}
\usepackage{soul}
\usepackage{url}
\usepackage[hidelinks]{hyperref}
\usepackage[utf8]{inputenc}
\usepackage[small]{caption}
\usepackage{graphicx}
\usepackage{amsthm}
\usepackage{booktabs}
\usepackage{algorithm}
\usepackage{amsmath}
\usepackage{amssymb}
\usepackage{algpseudocode} 
\usepackage{amsfonts}
\usepackage{hyperref}

\title{IRL with Partial Observations using the Principle of Uncertain Maximum Entropy}

\author{
Kenneth Bogert$^1$\footnote{Contact Author}\and
Yikang Gui$^2$\And
Prashant Doshi$^2$\\
\affiliations
$^1$University of North Carolina Asheville\\
$^2$THINC Lab, University of Georgia\\
\emails
kbogert@unca.edu,
Yikang.Gui@uga.edu,
pdoshi@uga.edu
}
\begin{document}

\maketitle
\begin{abstract}
The principle of maximum entropy is a broadly applicable technique for computing a distribution with the least amount of information possible while constrained to match empirically estimated feature expectations. However, in many real-world applications that use noisy sensors computing the feature expectations may be challenging due to partial observation of the relevant model variables. For example, a robot performing apprenticeship learning may lose sight of the agent it is learning from due to environmental occlusion. We show that in generalizing the principle of maximum entropy to these types of scenarios we unavoidably introduce a dependency on the learned model to the empirical feature expectations. We introduce the principle of uncertain maximum entropy and present an expectation-maximization based solution generalized from the principle of latent maximum entropy. Finally, we experimentally demonstrate the improved robustness to noisy data offered by our technique in a maximum causal entropy inverse reinforcement learning domain.
\end{abstract}

\section{Introduction}

The principle of maximum entropy is a technique for finding a distribution over random variable(s) $X$ that contains the least amount of information while still conforming to constraints. It has existed in various forms since the early 20$^{th}$ century but was formalized by Jaynes~\cite{Jaynes_MaxEnt}.  In a commonly encountered form, the constraints consist of matching sufficient statistics (e.g., feature expectations) under the maximum entropy model being learned and those observed empirically. 

However, in many cases the feature expectations are not directly observable. It could be the case that the model contains hidden variables, that some data is missing or corrupted by noise, or that $X$ is only partially observable using existing sensors or processes. Such scenarios are common when attempting to use maximum entropy (MaxEnt) models to perform inference using real-world data.

As an example, let us take a simple natural language processing model. Using the principle of maximum entropy, each $X$ may be a word in a vocabulary $\mathbb{X}$, and we wish to obtain a model that matches the empirical distribution of words in a given document, $\Tilde{Pr}(X)$, as given by the expectation of some interesting features $\phi_k(X)$.

However, if the data gathered is an audio recording of spoken words then the words themselves are never directly observed.  Instead, we may extract observations $\omega \in \Omega$ from the recording that only partially reveals the words being spoken. For instance, suppose $\Omega$ corresponds to phonemes. $Pr(\omega|X)$, the probability of hearing a phoneme for a given word, will be non-deterministic as different dialects and accents pronounce the same word in different ways.  Further, a poor quality voice recording may cause uncertainty in the phoneme being spoken, forcing the use of a more general $\omega$ to correctly model the data received. 

We make the following contributions in this work.
\begin{enumerate}
\item  We introduce the principle of uncertain maximum entropy as a new non-linear program and analyze it.
\item  We show a solution to the non-linear program using expectation-maximization, which has been previously used to solve the principle of latent maximum entropy~\cite{Wang2002,Bogert16:Expectation}.
\item  We apply the new principle to the the domain of inverse reinforcement learning (IRL)~\cite{Ng00:Algorithms} by generalizing the maximum causal entropy IRL algorithm~\cite{Ziebart2010} to observation noise. We experimentally demonstrate the efficacy of our new technique in a toy IRL application.
\end{enumerate}

\section{Background}

We briefly review the principles of maximum entropy and latent maximum entropy, and discuss their use in IRL. 

\subsection{Principle of Maximum Entropy}

The principle of maximum entropy may be expressed as a non-linear program.
\begin{small}
\begin{align}
&  \max \limits_{\Delta} \left( -\sum\nolimits_{X \in \mathbb{X}} Pr(X)~ \log~Pr(X) \right )\nonumber\\
&  \mbox{{\bf subject to}} \nonumber\\
& \sum \nolimits_{X \in \mathbb{X}} Pr(X) = 1 \nonumber\\
&  \sum \nolimits_{X \in \mathbb{X}} Pr(X) \phi_k(X)  = 
\hat{\phi}_k  ~~~~~~ \forall k 
\label{eq:max_ent}
\end{align}
\end{small}

Where $\hat{\phi}_k ~=~ \sum \nolimits_{X \in \mathbb{X}} \Tilde{Pr}(X) \phi_k(X)$ are the empirical feature expectations .
Notably, this program is known to be convex which provides a number of benefits.  Particularly relevant is that we may find a closed-form definition of $Pr(X)$, and solving the problem's dual is guaranteed to also solve the primal problem~\cite{Boyd2002}.  The solution is a log linear or Boltzmann distribution over the model elements,
$Pr(X) ~=~ \frac{e^{\sum\limits_{k=1}^K \lambda_k \phi_k(X)}}{Z(\lambda)}$
Where $Z(\lambda) ~=~ e^{-1}e^\eta ~=~ \sum\limits_{X' \in \mathbb{X}} e^{\sum\limits_{k=1}^K \lambda_k \phi_k(X')}$.

\subsection{Principle of Latent Maximum Entropy}

First presented in \cite{wang2001latent}, the principle of latent maximum entropy generalizes the principle of maximum entropy to models with hidden variables that are never empirically observed.

Split each $X$ into $Y$ and $Z$.  $Y$ is the component of X that is perfectly observed, $Z$ is perfectly un-observed and completes $Y$.  Thus, $X = Y \cup Z$ and $Pr(X) ~=~ Pr(Y, Z)$.  Latent maximum entropy corrects for the hidden portion of $X$ in the empirical data by summing over all $Z \in Z_Y$, which is every way of completing a given $Y$ to arrive at a $X$.  Its non-linear program is identical to equation~\ref{eq:max_ent} except the empirical feature expectations $\hat{\phi}_k ~=~ \sum \limits_{Y \in \mathbb{Y}} \Tilde{Pr}(Y) \sum\limits_{Z \in Z_Y} Pr(Z | Y) \phi_k(X)$.


Since  $Pr(Z|Y) ~=~ Pr(X) / Pr(Y)$, the right side of the feature expectations contains a dependency on the model being learned, meaning the program is no longer convex and only an approximate solution can be found if we still desire a log-linear model for $Pr(X)$.  This leads to an expectation-maximization solution, the resulting algorithm fills in the missing portions of the data with the complete model given current parameters, which are successively improved by EM.  To our knowledge \cite{wang2001latent} is the first to apply EM to the principle of maximum entropy to account for incomplete data.  

As the methodology and arguments used in this work are very similar to those  in \cite{Wang2012} the reader is encouraged to review \cite{Wang2012} for more background and discussion.

\subsection{Inverse Reinforcement Learning}

A Markov Decision Process \cite{puterman2014markov} is defined as the tuple $<S, A, T, R, \gamma, S_0>$, where $S$ is a set of states, $A$ is a set of actions, $T$ is a function describing transition probabilities, $Pr(S | S, A)$, $R: S \times A \rightarrow \mathbb{R}$ is a reward function mapping states and actions to a real number, $\gamma$ is a discount rate and $S_0$ is the initial state distribution.  MDP's model a rational agent that is Markovian, ie, its state variable completely describes all relevant information such that it does not need to consider its history when making decisions.  A MDP is solved by using Reinforcement Learning to arrive at  optimal behavior which may be expressed as a policy $\Pi = Pr(A|S)$  that describes the action to choose for every state.

IRL reverses this process.  First presented by \cite{Ng00:Algorithms}, we begin with a MDP missing its reward function and the behavior of an assumed optimal agent.  We are tasked with finding a $R$ such that the agent's behavior is optimal in the resulting complete MDP.  Applications of IRL include Apprenticeship Learning \cite{Abbeel04:Apprenticeship}, wherein a learner agent wishes to perform a task as well as an expert agent.  The learner utilizes IRL to recover a reward function that describes the expert's behavior, and uses the learned rewards to complete its own MDP.  By solving this MDP the learner arrives at a policy that may perform the same task in a different way, such as if it is physically dissimilar from the expert. 

Unfortunately, IRL is ill-posed as there exists an infinite number of reward functions under which a given behavior is optimal \cite{Ng00:Algorithms}.  This makes evaluating the performance of IRL algorithms complicated, as directly comparing learned reward functions is not fruitful.  In our experiments, we use a popular measure, {\em inverse learning error} (ILE)~\cite{Choi11:Inverse} of a learned reward $R^L$ obtained as ILE $= || V^{\pi^E} - V^{\pi^L}||_1$, where $V^{\pi^E}$ is the value function computed using the true reward and the expert's optimal policy, $V^{\pi^L}$ is the value function using the true rewards and the optimal policy for the \textbf{learned} reward.

To address the ill-posedness of IRL, \cite{Ziebart08:Maximum} introduced Maximum Entropy IRL.  This algorithm makes use of the principle of maximum entropy by choosing $\mathbb{X}$ to be the set of all possible MDP trajectories of the target agent.  By assuming $R(S,A)$ is a linear combination of $\phi_k(S,A)$ feature functions MaxEntIRL finds the one set of weights that results in the maximum entropy over $\mathbb{X}$ while matching empirical feature expectations.  Later, \cite{ZiebartBD_compare_MMP} presented the MaxCausalEntIRL algorithm which extended MaxEntIRL to non-deterministic MDPs. \cite{Bogert16:Expectation} extended MaxEntIRL to scenarios with missing data by employing the principle of latent maximum entropy.  The resulting algorithm, HiddenDataEM is restricted to scenarios with perfectly unknown sections of an agent's trajectory, due to occlusion of the environment for instance.  It requires perfect knowledge of all other portions of the trajectory and is therefore unable to make use of any available partial information.

\section{Uncertain Maximum Entropy}

For our application domain of IRL with noisy real-world data we require an algorithm that both accounts for the ill-posedness of the problem and can make use of imperfect empirical data from noisy sensors. We begin by generalizing the principle of latent maximum entropy (and by extension, the principle of maximum entropy) to uncertain data, which we call the {\em principle of uncertain maximum entropy}. 

Assume we want a maximum entropy model of some variables $X \in \mathbb{X}$ given we have observations $\omega \in \Omega$.  Critically, we desire that the model does not include $\omega$ as the observations themselves will pertain solely to the data gathering technique used to produce it, not the elements or model being observed. Additionally, we assume the existence of a static observation function $Pr(\omega | X)$. Let $\tilde{Pr}(\omega)$ be the empirical distribution of observations obtained from data. Our new non-linear program is:

\begin{align}
&  \max \limits_{\Delta} \left( -\sum\nolimits_{X \in \mathbb{X}} Pr(X)~ \log~Pr(X) \right )\nonumber\\
&  \mbox{{\bf subject to}} \nonumber\\
& \sum \nolimits_{X \in \mathbb{X}} Pr(X) = 1 \nonumber\\
&  \sum \limits_{X \in \mathbb{X}} Pr(X) \phi_k(X)  = \sum \limits_{\omega \in \Omega} \Tilde{Pr}(\omega) \sum \limits_X Pr(X | \omega) ~\phi_k(X) ~~~ \forall k 
\label{eq:uncertain_latent_max_ent}
\end{align}

To solve Eq~\ref{eq:uncertain_latent_max_ent}, we first take the Lagrangian.
\begin{small}
\begin{align}
&\mathcal{L}(X, \Omega, \lambda, \eta) ~=~ -\sum\nolimits_{X \in \mathbb{X}} Pr(X)~\log~Pr(X) + \nonumber \\
&\eta \left ( \sum \nolimits_{X \in \mathbb{X}} Pr(X) - 1 \right ) + \nonumber \\
&\sum\limits_{k=1}^K \lambda_k \left ( \sum \limits_{X \in \mathbb{X}} Pr(X) \phi_k(X) - \sum \limits_{\omega \in \Omega} \Tilde{Pr}(\omega) \sum \limits_X Pr(X | \omega) ~\phi_k(X) \right ). 
\label{eq:lagrange}
\end{align}
\end{small}

Next, we find Lagrangian's gradient so that we can set it to zero and attempt to solve for $Pr(X)$.
\begin{small}
\begin{align}
&{\frac{\partial \mathcal{L}(X, \Omega, \lambda, \eta)} {\partial Pr(X)}} ~=~ \log Pr(X) - 1 + \eta + \sum\limits_{k=1}^K \lambda_k \phi_k(X) ~- \nonumber \\
&\sum\limits_{k=1}^K \lambda_k \sum \limits_{\omega \in \Omega} \Tilde{Pr}(\omega) \left ( \phi_k(X)  {\frac{ Pr(\omega | X) Pr(\omega) - Pr(\omega | X) ^2 Pr(X)} { Pr(\omega)^2}} \right ).
\label{eq:lagrange_gradient}
\end{align}
\end{small}

Unfortunately, the existence of $Pr(X | \omega)$ on the right side of the constraints causes the gradient to be non-linear in $Pr(X)$. However, Wang et al.~(\citeyear{Wang2012}) notes that we may approximate the gradient, which then allows $Pr(X)$ to be log-linear. In other words:
\begin{align}
\frac{\partial \mathcal{L}(X, \Omega, \lambda, \eta)} {\partial Pr(X)} & ~\approx~ -\log Pr(X) - 1 + \eta + \sum\limits_{k=1}^K \lambda_k \phi_k(X) \nonumber \\
0 & ~\approx~ -\log Pr(X) - 1 + \eta + \sum\limits_{k=1}^K \lambda_k \phi_k(X) \nonumber \\
\mbox{which gives,} \nonumber\\
Pr(X) &~\approx~ \frac{e^{\sum\limits_{k=1}^K \lambda_k \phi_k(X)}} {Z(\lambda)}
\label{eq:log-linear}
\end{align}

Now we enter our approximation back into the Lagrangian to arrive at an approximate dual:
\begin{small}
\begin{equation}
\mathcal{L}_{dual}(\lambda) ~\approx~ \log Z(\lambda) - \sum\limits_{k=1}^K \lambda_k \sum \limits_{\omega \in \Omega} \Tilde{Pr}(\omega) \sum \limits_X Pr(X | \omega) \phi_k(X).
\label{eq:dual}
\end{equation}
\end{small}
We would now try to find the dual's gradient and use it to minimize the dual.  Unfortunately the presence of $Pr(X|\omega)$ still admits no closed form solution in general.  We will instead employ another technique to minimize it as we show next.

\subsection{EM Solution}

Inspired by Wang et al.~(\citeyear{Wang2012}) the constraints of Eq.~\ref{eq:uncertain_latent_max_ent} can be satisfied by locally maximizing the likelihood of the observed data. One method is to derive an EM algorithm utilizing our log-linear model. We begin with the log likelihood of all the observations, note subscripts indicate a distribution is parameterized with the respective variable:
\begin{small}
\begin{align}
L(\lambda) ~=~& \sum\limits_{\omega \in \Omega} \Tilde{Pr}(\omega) ~\log~ Pr(\omega) \nonumber \\
 ~=~& \sum\limits_{\omega \in \Omega} \Tilde{Pr}(\omega) ~\log~ \sum\limits_{X \in \mathbb{X}} Pr_\lambda(\omega, X) \nonumber\\
 ~=~& \sum\limits_{\omega \in \Omega} \Tilde{Pr}(\omega) ~\log~ \sum\limits_{X \in \mathbb{X}} \frac{ Pr_\lambda(\omega, X)} {Pr_{\lambda'} (X | \omega)} Pr_{\lambda'} (X | \omega) \nonumber\\
 ~\geq~& \sum\limits_{\omega \in \Omega} \Tilde{Pr}(\omega) \sum\limits_{X \in \mathbb{X}} Pr_{\lambda'} (X | \omega) \log \frac{ Pr_\lambda(\omega, X)}{Pr_{\lambda'}(X | \omega) } \nonumber\\
 ~=~& \sum\limits_{\omega \in \Omega} \Tilde{Pr}(\omega) \sum\limits_{X \in \mathbb{X}} Pr_{\lambda'} (X | \omega) \log Pr_\lambda(\omega, X) - \nonumber \\
 &\sum\limits_{\omega \in \Omega} \Tilde{Pr}(\omega) \sum\limits_{X \in \mathbb{X}} Pr_{\lambda'} (X | \omega) \log  Pr_{\lambda'}(X | \omega) \nonumber\\
 ~=~& \sum\limits_{\omega \in \Omega} \Tilde{Pr}(\omega) \sum\limits_{X \in \mathbb{X}} Pr_{\lambda'} (X | \omega) \log Pr_\lambda(\omega| X) Pr_\lambda(X) \nonumber \\
& + H(\lambda') \nonumber\\
 ~=~& \sum\limits_{\omega \in \Omega} \Tilde{Pr}(\omega) \sum\limits_{X \in \mathbb{X}} Pr_{\lambda'} (X | \omega) \log Pr_\lambda(\omega| X) \nonumber \\
&+ \sum\limits_{\omega \in \Omega} \Tilde{Pr}(\omega) \sum\limits_{X \in \mathbb{X}} Pr_{\lambda'} (X | \omega) \log Pr_\lambda(X) + H(\lambda') \nonumber \\
\label{eq:EM_U_star}
 ~=~& \sum\limits_{\omega \in \Omega} \Tilde{Pr}(\omega) \sum\limits_{X \in \mathbb{X}} Pr_{\lambda'} (X | \omega) \log Pr(\omega| X) \nonumber \\
 &+ Q(\lambda, \lambda') + H(\lambda')\\
 ~=~& U^*(\lambda') + Q(\lambda, \lambda') + H(\lambda').
\end{align}
\end{small}

Equation~\ref{eq:EM_U_star} follows because the observation function $Pr(\omega | X)$ does not depend on $\lambda$.  This leaves $Q(\lambda, \lambda')$ as the only function which depends on $\lambda$.  The EM algorithm proceeds by maximizing $Q$, and upon convergence $\lambda = \lambda'$, at which point the likelihood of the data is at a local maximum.

For the sake of completeness, we note that $H(\lambda')$ is the conditional entropy of the latent variables and $U^*(\lambda')$ is the expected log observations. These impact the overall data likelihood only, not the model solution because the observations are not included in the model. We now substitute the log-linear model of $Pr(X)$ (Eq.~\ref{eq:log-linear}) into $Q(\lambda, \lambda')$:

\begin{small}
\begin{align}
&Q(\lambda, \lambda') ~=~ \sum\limits_{\omega \in \Omega} \Tilde{Pr}(\omega) \sum\limits_{X \in \mathbb{X}} Pr_{\lambda'} (X | \omega) ~\log Pr_\lambda(X) \nonumber \\
& ~=~ \sum\limits_{\omega \in \Omega} \Tilde{Pr}(\omega) \sum\limits_{X \in \mathbb{X}} Pr_{\lambda'} (X | \omega) \left (  \sum\limits_{k=1}^K \lambda_k \phi_k(X) - ~\log Z(\lambda) \right ) \nonumber \\
& ~=~ - \log Z(\lambda) + \sum\limits_{k=1}^K \lambda_k \sum\limits_{\omega \in \Omega} \Tilde{Pr}(\omega) \sum\limits_{X \in \mathbb{X}} Pr_{\lambda'} (X | \omega)  \phi_k(X). 
\label{eq:em_finished}
\end{align}
\end{small}

Notice that Eq.~\ref{eq:em_finished} is similar to Eq.~\ref{eq:dual}. One important difference is that Eq.~\ref{eq:em_finished} is easier to solve as $Pr(X |\omega)$ depends on $\lambda'$ and not $\lambda$.  In fact, maximizing $Q(\lambda, \lambda')$ is equivalent to solving the following program by minimizing its dual, Eq.~\ref{eq:dual}:
\begin{align}
&  \max \limits_{\Delta} \left( -\sum\nolimits_{X \in \mathbb{X}} Pr_\lambda(X)~ \log~Pr_\lambda(X) \right )\nonumber\\
&  \mbox{{\bf subject to}} \nonumber\\
& \sum \nolimits_{X \in \mathbb{X}} Pr_\lambda(X) = 1 \nonumber\\
&  \sum \limits_{X \in \mathbb{X}} Pr_\lambda(X) \phi_k(X)  = \sum \limits_{\omega \in \Omega} \Tilde{Pr}(\omega) \sum \limits_X Pr_{\lambda'}(X | \omega) ~\phi_k(X) ~~~ \forall k 
\label{eq:uncertain_latent_max_ent_em}
\end{align}
which equals Eq.~\ref{eq:uncertain_latent_max_ent} at convergence.  We now arrive at the following EM algorithm:

\begin{algorithm}
	\caption{uMaxEnt} 
	\begin{algorithmic}[1]
\State\emph{Start:} Initialize $\lambda'$
\State\emph{E Step:}  Using  $\lambda'$, compute: \Statex$\hat{\phi}_k ~=~ \sum \nolimits_{\omega \in \Omega} \Tilde{Pr}(\omega) \sum \nolimits_X Pr_{\lambda'}(X | \omega) ~\phi_k(X)$
\State\emph{M Step:} Solve Equation~\ref{eq:max_ent}'s program to arrive at a new $\lambda$. 
\State Set $\lambda' = \lambda$
\State\emph{Repeat:} Until $\lambda$ converges 
\end{algorithmic}
\end{algorithm}

\section{Analysis of New Principle}

To show that the principle of uncertain maximum entropy generalizes the principle of maximum entropy, let $\Omega$ equal to $\mathbb{X}$.  Then $\Tilde{Pr}(\Omega) = \Tilde{Pr}(X)$ and $\sum\limits_X Pr(X|\omega) = \sum\limits_{X'} Pr(X'|X)$. Hence, we recover Eq.~\ref{eq:uncertain_latent_max_ent}'s constraint terms.  Similarly, the principle of uncertain maximum entropy generalizes the principle of latent maximum entropy~\cite{Wang2002} by allowing $\Omega = \mathbb{Y}$, then $\Tilde{Pr}(\Omega) = \Tilde{Pr}(Y)$ and $\sum\limits_X Pr(X|\omega) = \sum\limits_{Z \in Z_Y} Pr(Z|Y)$ since $X = Y \cup Z$.

Additionally, in both cases the generalization holds with an arbitrary $\Omega$ so long as the following conditions hold: 
$\{ Pr(X | \omega) \in \{0, 1\} ~\forall~ X, \omega$ and $\exists ~\omega ~\ni~ Pr(X | \omega) ~=~ 1 ~\forall~ X \}$ 
and 
$\{ Pr(Y | \omega) \in \{0, 1\} ~\forall~ Y, \omega$ and $\exists ~\omega ~\ni~ Pr(Y | \omega) ~=~ 1 ~\forall~ Y \}$, respectively.  

One valuable property for statistical inference models is the ability to guarantee correctness in the infinite limit of available data by making use of the law of large numbers.  For linear and non-linear programs, this is satisfied by showing that the constraints hold true as the empirical distribution of data converges to the true distribution.  Next, we show that the principle of uncertain maximum entropy exhibits this attribute, while the principle of maximum entropy does not except in special cases. 

Notice that $Pr(X | \omega) ~=~ \frac{ Pr(\omega | X) Pr(X) } { Pr(\omega)  }$ and therefore in the infinite limit of data where $\Tilde{Pr}(\omega) ~=~ Pr(\omega)$ the constraints of are satisfied as:
\begin{align}
 \sum \nolimits_{X \in \mathbb{X}} &Pr(X) \phi_k(X)  \nonumber \\
 &= \sum \nolimits_{\omega \in \Omega} Pr(\omega) \sum \nolimits_X Pr(X | \omega) ~\phi_k(X) \nonumber \\
 & = \sum \nolimits_{\omega \in \Omega} Pr(\omega) \sum \nolimits_X \frac{ Pr(\omega | X) Pr(X) }{ Pr(\omega)  } ~\phi_k(X) \nonumber \\
 & = \sum \nolimits_{\omega \in \Omega} \sum \nolimits_X Pr(\omega | X) Pr(X) ~\phi_k(X) \nonumber \\
 & = \sum \nolimits_X  Pr(X) ~\phi_k(X) \sum \nolimits_{\omega \in \Omega} Pr(\omega | X) \nonumber \\
 & = \sum \nolimits_X  Pr(X) ~\phi_k(X).
\end{align}

To use the principle of maximum entropy with imperfect data the observations must be converted into the model elements.  For instance, in statistical physics, where the technique was first developed, the observations themselves are the result of a large number of smaller elements, such as measuring the temperature of a gas.  So long as we constrain the model with such statistical moments, we may safely use the principle of maximum entropy.

In general, however, we must accept some amount of error in the observation-to-model-element conversion.  As a result in the infinite limit of data $\tilde{Pr}(X) \neq Pr(X)$ and therefore the equality constraints do not hold.  The effects of this are ad-hoc, for instance errors in continuous data may be bounded to within an acceptable tolerance while categorization errors in discrete models may result in unbounded error.  This makes it difficult to use the technique with large data sets or in automated inference applications with noisy sensor data.

To illustrate, suppose we choose the most likely $X$ for an observed $\omega$, or in other words, each $X$'s probability is the sum of the observed $\omega$'s probabilities for which it is the most likely element. In the limit, $\tilde{Pr}(X) =  \sum\limits_{\omega } Pr(\omega) \delta ( \underset{X'}{\mathrm{argmax}}  ~Pr(X' | \omega) , X)$  where $\delta$ is the Kronecker delta.  This distribution does not equal $\sum\limits_{\omega } Pr(\omega) Pr(X | \omega)$, the true $Pr(X)$.

\begin{figure}[ht]
	\centering
	\includegraphics[width=0.45\textwidth]{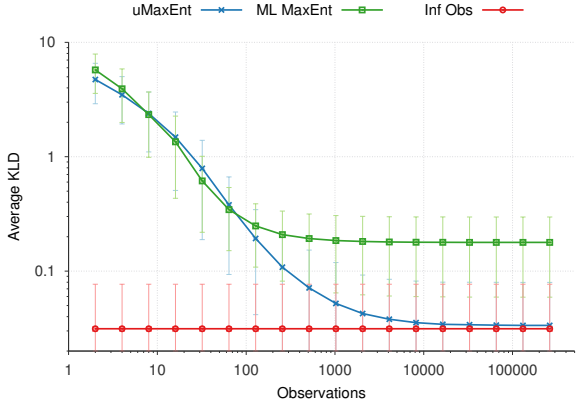}
	\caption{Average KLD between the learned distribution over $\mathbb{X}$ and the true distribution as observations increase in randomly generated uncertain maximum entropy programs.  Note log scale on both axis.  Error bars are standard deviation, each point the result of 1100 runs.}
	\label{fig:analysis}
	\vspace{-0.1in}
\end{figure}

To demonstrate empirically, we generated a large number of random maximum entropy programs of random size with random observation functions.  For each we generate a number of observations as well as their corresponding most-likely $X$ and use these in our new technique and principle of maximum entropy, respectively.  Figure~\ref{fig:analysis} shows the Kullback–Leibler divergence of the resulting learned models to the true models as the number of observations provided increases.  Notice that \textbf{uMaxEnt} continues improving as more observations are given until it reaches the performance of the the best case control \textbf{Inf Obs} (where $\tilde{Pr}(\omega)$ is set to $Pr(\omega)$.  Meanwhile, \textbf{ML MaxEnt} converges at a higher KLD and no additional data improves its performance.

\section{Experiment}

\begin{figure}[h]
	\centering
	\includegraphics[width=0.2\textwidth]{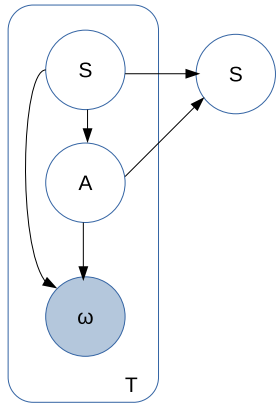}
	\caption{Plate diagram of a partially observed MDP-based agent.  Only $\omega$ is observed each timestep.  Note $Pr(A | S)$, the agent's policy, which is unknown and must be learned to compute the complete distribution. }
	\label{fig:hmm}
	\vspace{-0.1in}
\end{figure}

To evaluate the performance of our new technique in our inverse reinforcement learning application domain, we implemented a new algorithm, \textbf{uMaxCausalEntIRL}, by generalizing MaxCausalEntIRL using the principle of uncertain maximum entropy.  Each $X$ becomes a tuple composed of elements $<s, a>_t$ corresponding to one trajectory of the MDP.  The empirical data of the agent being learned from may now be thought of as a specialized Hidden-Markov model, as shown in figure~\ref{fig:hmm}.  In this view we concretely see the empirical data's dependency on the model being learned as the distribution over all states and actions over time given the observations may not be calculated due to the missing distribution $Pr(A | S)$.  This distribution, the agent's policy, is only available after our EM-based algorithm is complete. 

\begin{figure}[h]
	\centering
	\includegraphics[width=0.3\textwidth]{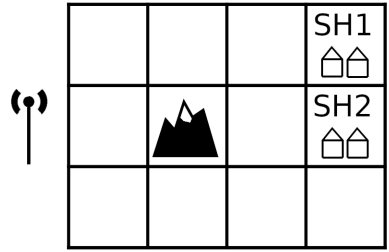}
	\caption{Illustration of the fugitive scenario.  The fugitive is attempting to reach safe house 1 (SH1) and may start in any state.  Note the location of the radio tower on the left and the impassible mountain range.}
	\label{fig:fugitive}
	\vspace{-0.1in}
\end{figure}

Our scenario is that of a fugitive being unknowingly tracked by a radio tower that is listening for pings coming from a tracker device hidden on them.  We attempt to discern the fugitive movement's within a space including identifying a safe house.  Complicating matters the tracking is imperfect, becoming less accurate the farther away the fugitive is from the tower.  Further, it's possible the tower may miss a transmission and there is a mountain range that greatly increases the probability of missing transmissions coming from behind it.  We illustrate this scenario in figure~\ref{fig:fugitive}.

We model the fugitive's movement using a MDP with 11 states and four actions corresponding to moving in the four cardinal directions.  Movement is uncertain, as the fugitive may attempt to move in a direction but be unable to during a given timestep.  This is modeled as an action having its intended movement effect 90\% of the time, with the remaining probability mass given to the other actions' results (invalid movements cause the fugitive to remain in the same state).  There are two potential safe houses as marked on the map, with the correct one being SH1, and they are attempting to reach it.  The fugitive may start from any state. 

\begin{figure}[!ht]
	\centering
	\includegraphics[width=.45\textwidth]{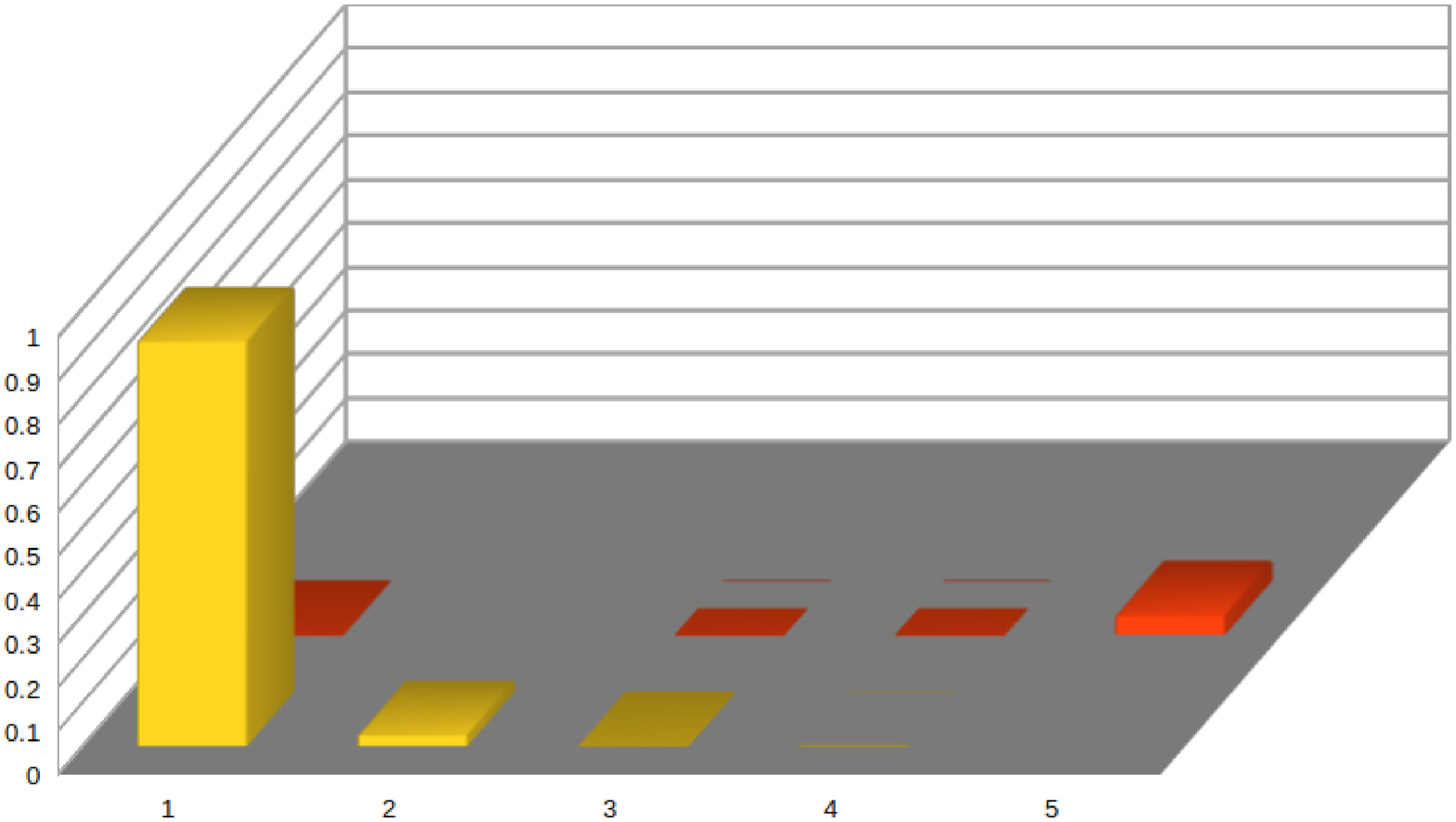}
	\includegraphics[width=.45\textwidth]{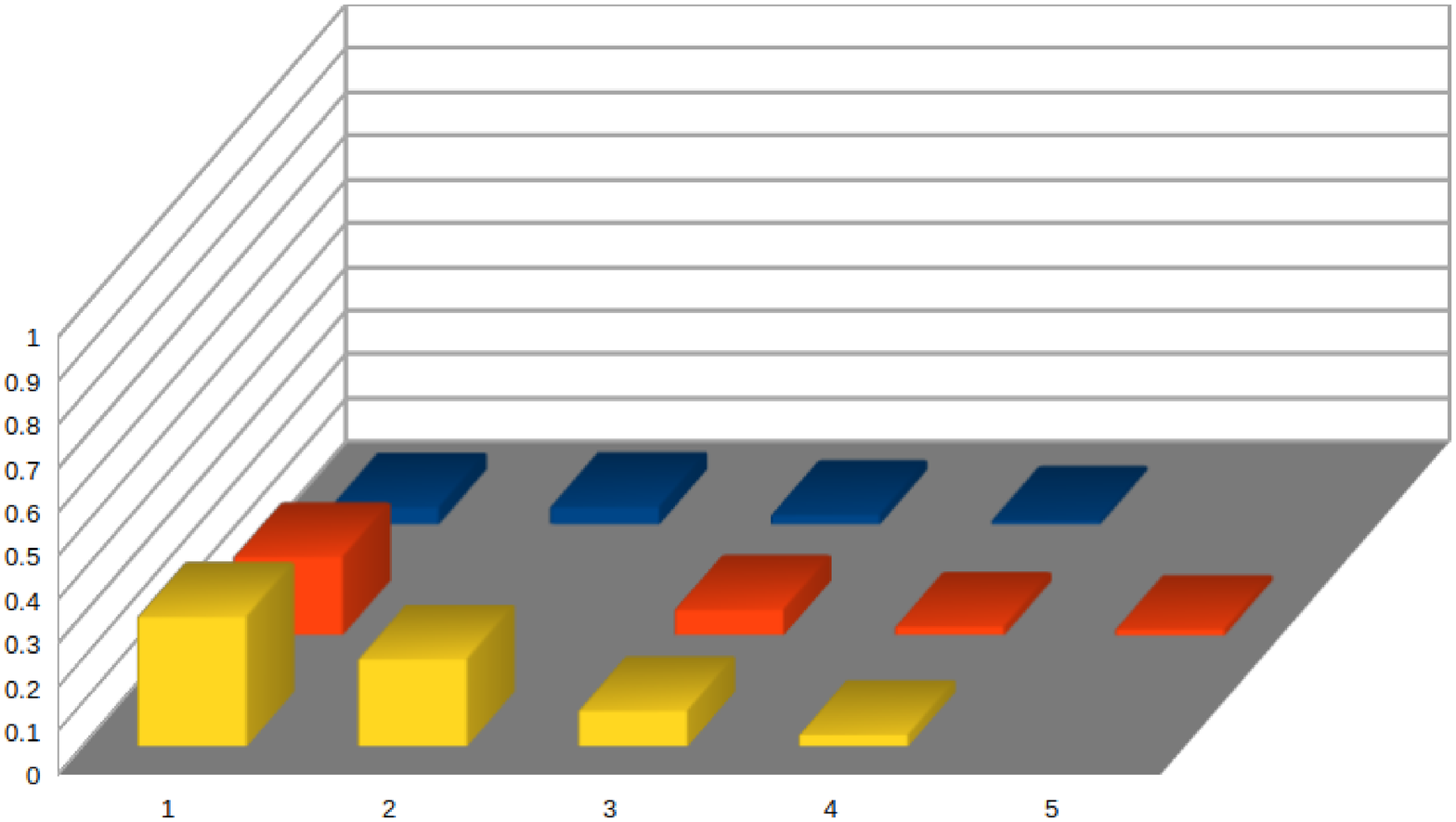}
	\caption{\small Illustration of $Pr(\Omega | s)$ for typical low (Top) and high (Bottom) observation error settings.  The location of the fugitive in both cases is the bottom left.  Bars are arranged such that each $\omega_s$ is over its corresponding state.  $\omega_f$ is shown in the last column on the right.}
	\label{fig:fugitiveobs}
	\vspace{-0.1in}
\end{figure}

We model the tower's observations using 12 $\omega$; 11 corresponding each to one state $\omega_s$ that represents the most likely position of the fugitive being that state, and one $\omega_f$ to represent "no ping received".  We vary the amount of noise in the observations by changing the standard deviation of $Pr(\omega_s | s)$, a "low" setting makes the "correct" observation (and those nearby) much more likely, while "high" spreads probability mass among most of the observations.  We illustrate an example $Pr(\Omega | s)$ with these two standard deviation settings in figure~\ref{fig:fugitiveobs}.

We develop a number of controls to better illustrate our new technique's performance.  \textbf{TRUE} is the best case scenario where we record the fugitive's true trajectories and use MaxCausalEntIRL.  \textbf{ML} is an unrealistic worst-case procedure where we simply record the most likely state for each received observation, and in the case of a "missed ping" observation this absurdly records SH2.  A more realistic procedure, \textbf{WOERR} discards the "missed ping" observations to create missing data.  Both of these controls use MaxCausalEntIRL on their resulting data sets.  Finally, \textbf{cHiddenDataEM}, a simple adaptation of HiddenDataEM to use MaxCausalEntIRL, performs HiddenDataEM on the dataset produced by WOERR to account for the missing data.

\begin{figure}[!ht]
	\centering
	\includegraphics[width=.45\textwidth]{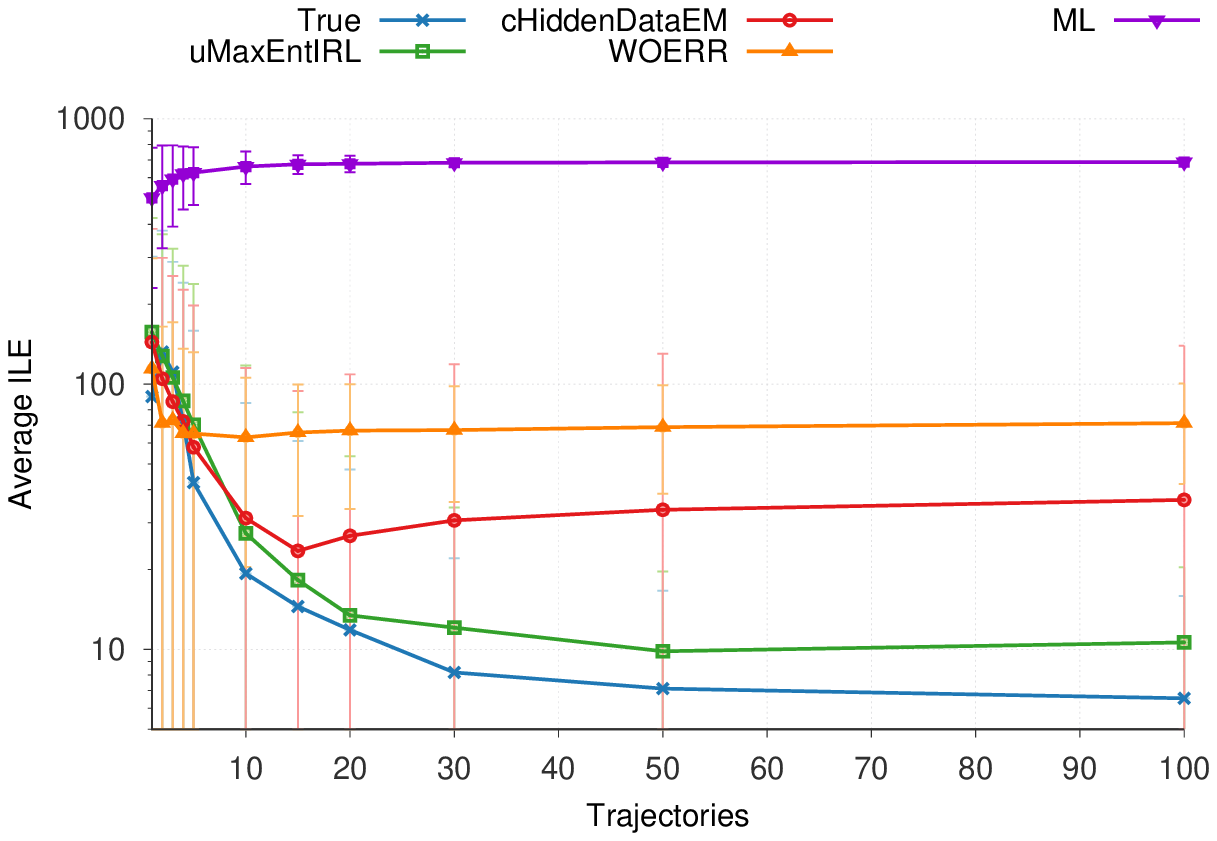}
	\includegraphics[width=.45\textwidth]{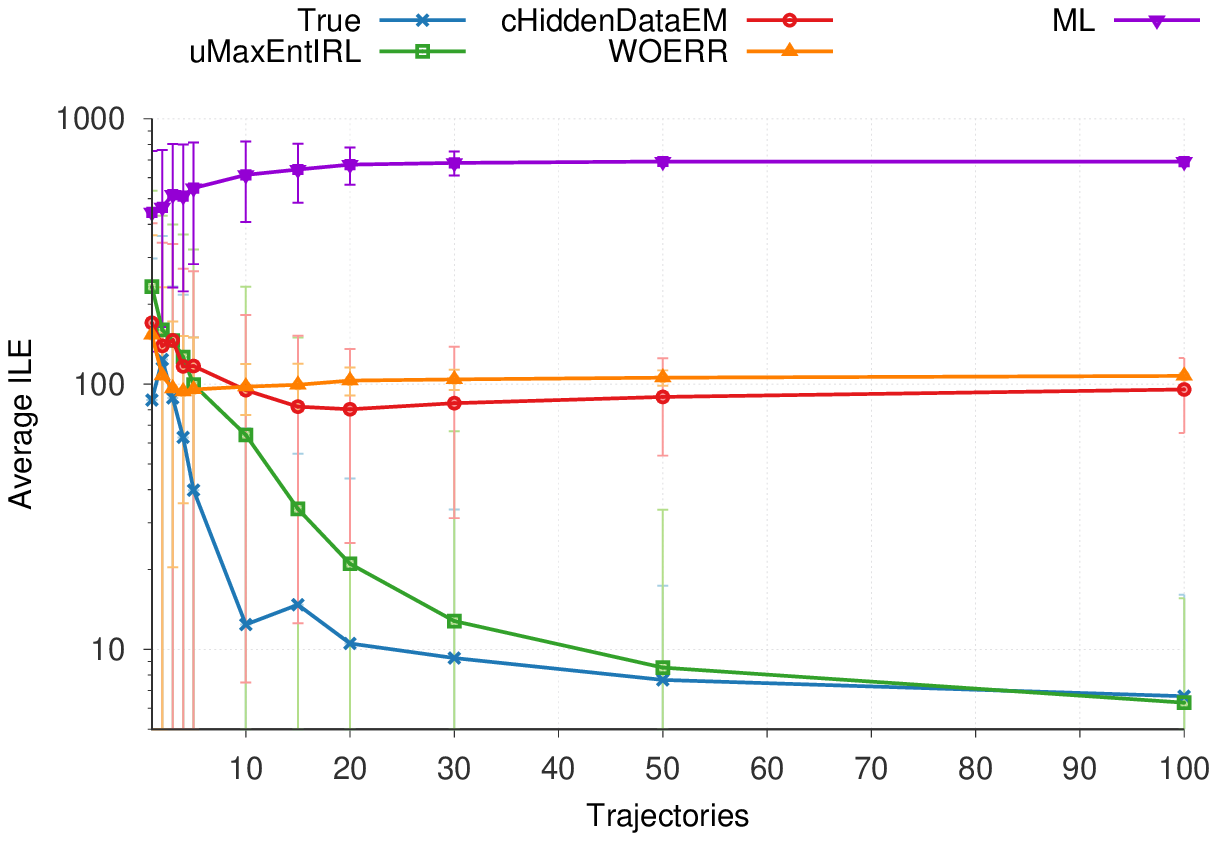}
	\caption{\small Average ILE achieved by various algorithms in the fugitive scenario with low (Top) and high (Bottom) observation error settings.  Note log scale on y axis.  Error bars are standard deviation.  Points are the result of 800 (low) and 160 (high) trials}
	\label{fig:fugitivegraph}
	\vspace{-0.1in}
\end{figure}

We present our results in figure~\ref{fig:fugitivegraph} showing the average ILE achieved by each algorithm as the number of trajectories increases.  Notice that in the low error setting cHiddenDataEM improves at first but as more trajectories are provided it stops improving at a relatively high error, ML fails to improve at all and WOERR stops improving after a very short number of trajectories.  In the high error setting the control algorithms show minimal improvement after the first few trajectories. Only uMaxCausalEntIRL is able to learn accurately in both error settings. 

\section{Related Work}

To our knowledge Maximum Entropy models utilizing imperfect data have received comparatively little attention.  By contrast, applications with missing data are more common and may use latent maximum entropy to account for missing data, such as missing labels in a supervised learning scenario \cite{grandvalet2006entropy}, species distributions \cite{phillips2006maximum}, or language modeling \cite{wang2001latent}.  In the context of IRL missing data may be interpreted as occlusion of the agent being observed (\cite{Bogert14:Multi}, \cite{Bogert15:Toward}, \cite{Bogert18:Multi}).

IRL strongly motivates the user of partial observations due to the desire for deploying applications like apprenticeship learning in the real world.  \cite{Kitani2012} first introduce MaxEntIRL with partial observations, however, the learned model was not exploited to improve the state distribution leading to potentially incorrect results.  \cite{Shahryari2017} developed a MaxEntIRL with partial observations model based upon Latent Maximum Entropy that included the observations in the model.  But, the resulting model is difficult to interpret from a theoretical standpoint due to the joint distribution $Pr(X, \Omega)$ being log linear and not $Pr(X)$.  Recently, \cite{Unhelkar2019} presented IRL with hidden internal states but still required perfect observation of "external" states and actions.  

\section{Concluding Remarks}

The inability to make use of partial observations has been a significant obstacle to deploying apprenticeship learning in many real-world scenarios due to the rarity of error-free data.  In this work we present a new IRL technique, uMaxCausalEntIRL, based upon the principle of uncertain maximum entropy and demonstrate that it successfully generalizes MaxCausalEntIRL.  Future work will focus on further reduction of the engineering required by relaxing our algorithms requirements allowing the transition and observation functions to be unknown and estimated automatically from data, and on deploying our algorithm in an autonomous robotic apprenticeship learning scenario. 

\bibliographystyle{bst/named}
\bibliography{bydijcai22}

\end{document}